\documentclass{article}

\usepackage{amsmath,amsfonts,bm}

\def\eqref#1{equation~\ref{#1}}

\def\1{\bm{1}}

\def\ry{{\textnormal{y}}}
\def\rz{{\textnormal{z}}}

\def\rvx{{\mathbf{x}}}
\def\rvy{{\mathbf{y}}}
\def\rvz{{\mathbf{z}}}

\def\rmW{{\mathbf{W}}}

\def\vr{{\bm{r}}}

\def\vx{{\bm{x}}}
\def\vy{{\bm{y}}}
\def\vz{{\bm{z}}}

\def\mI{{\bm{I}}}

\def\mW{{\bm{W}}}

\DeclareMathAlphabet{\mathsfit}{\encodingdefault}{\sfdefault}{m}{sl}
\SetMathAlphabet{\mathsfit}{bold}{\encodingdefault}{\sfdefault}{bx}{n}

\DeclareMathOperator*{\argmax}{arg\,max}

\usepackage{microtype}
\usepackage{graphicx}
\usepackage{subfigure}
\usepackage{booktabs} 

\usepackage{hyperref}

\usepackage[accepted]{icml2020}

\usepackage{dsfont}

\begin{document}

\twocolumn[
\icmltitle{Being Bayesian about Categorical Probability}
\begin{icmlauthorlist}
\icmlauthor{Taejong Joo}{est}
\icmlauthor{Uijung Chung}{est}
\icmlauthor{Min-Gwan Seo}{est}
\end{icmlauthorlist}

\icmlaffiliation{est}{ESTsoft, Republic of Korea}

\icmlcorrespondingauthor{Taejong Joo}{tjoo@estsoft.com}

\icmlkeywords{Deep learning, Bayesian principle, Neural network, Softmax alternative, Variational inference}

\vskip 0.3in

]

\printAffiliationsAndNotice{}

\begin{abstract}
Neural networks utilize the softmax as a building block in classification tasks, which contains an overconfidence problem and lacks an uncertainty representation ability. As a Bayesian alternative to the softmax, we consider a random variable of a categorical probability over class labels. 
In this framework, the prior distribution explicitly models the presumed noise inherent in the observed label, which provides consistent gains in generalization performance in multiple challenging tasks. 
The proposed method inherits advantages of Bayesian approaches that achieve better uncertainty estimation and model calibration. Our method can be implemented as a plug-and-play loss function with negligible computational overhead compared to the softmax with the cross-entropy loss function.
\end{abstract}

\section{Introduction}
Softmax \citep{bridle1990probabilistic} is the de facto standard for post processing of logits of neural networks (NNs) for classification. 
When combined with the maximum likelihood objective, it enables efficient gradient computation with respect to logits and has achieved state-of-the-art performances on many benchmark datasets.
However, softmax lacks the ability to represent the uncertainty of predictions \citep{blundell2015weight,gal2016dropout} and has poorly calibrated behavior \citep{guo2017calibration}.
For instance, the NN with softmax can easily be fooled to confidently produce wrong outputs; when rotating digit 3, it will predict it as the digit 8 or 4 with high confidence \citep{louizos2017multiplicative}. 
Another concern of the softmax is its confident predictive behavior makes NNs to be subject to overfitting \citep{xie2016disturblabel, pereyra2017regularizing}. This issue raises the need for effective regularization techniques for improving generalization performance.

Bayesian NNs (BNNs; \citeauthor{mackay1992practical}, \citeyear{mackay1992practical}) can address the aforementioned issues of softmax. BNNs provide quantifiable measures of uncertainty such as predictive entropy and mutual information \citep{gal2016uncertainty} and enable automatic embodiment of Occam's razor \citep{mackay1995probable}. 
However, some practical obstacles have impeded the wide adoption of BNNs.
First, the intractable posterior inference in BNNs demands approximate methods such as variation inference (VI; \citeauthor{graves2011practical}, \citeyear{graves2011practical}; \citeauthor{blundell2015weight}, \citeyear{blundell2015weight}) and Monte Carlo (MC) dropout \citep{gal2016dropout}.
Even with such novel approximation methods, concerns arise regarding both the degree of approximation and the computational expensive posterior inference \citep{wu2019deterministic, osawa2019practical}. 
In addition, under extreme non-linearity between parameters and outputs in the NNs, determining a meaningful weight prior distribution is challenging \citep{sun2019functional}.
Last but not least, BNNs often require considerable modifications to existing baselines, or they result in performance degradation \citep{lakshminarayanan2017simple}. 

In this paper, we apply the Bayesian principle to construct the target distribution for learning classifiers. 
Specifically, we regard a categorical probability as a random variable, and construct the target distribution over the categorical probability by means of the Bayesian inference, which is approximated by NNs.
The resulting target distribution can be thought of as being regularized via the prior belief whose impact is controlled by the number of observations.
By considering only the random variable of categorical probability, the Bayesian principle can be efficiently adopted to existing deep learning building blocks without huge modifications.
Our extensive experiments show effectiveness of being Bayesian about the categorical probability in improving generalization performances, uncertainty estimation, and calibration property.

Our contributions can be summarized as follows: 
1) we show the importance of considering categorical probability as a random variable instead of being determined by the label;
2) we provide experimental results showing the usefulness of the Bayesian principle in improving generalization performance of large models on standard benchmark datasets, e.g., ResNext-101 on ImageNet;
3) we enable NNs to inherit the advantages of the Bayesian methods in better uncertainty representation and well-calibrated behavior with a negligible increase in computational complexity.

\begin{figure*}
    \centering
    \hfill
    \begin{subfigure} [Softmax cross-entropy loss \label{fig1_sm}]
     {\includegraphics[width=0.40\textwidth]{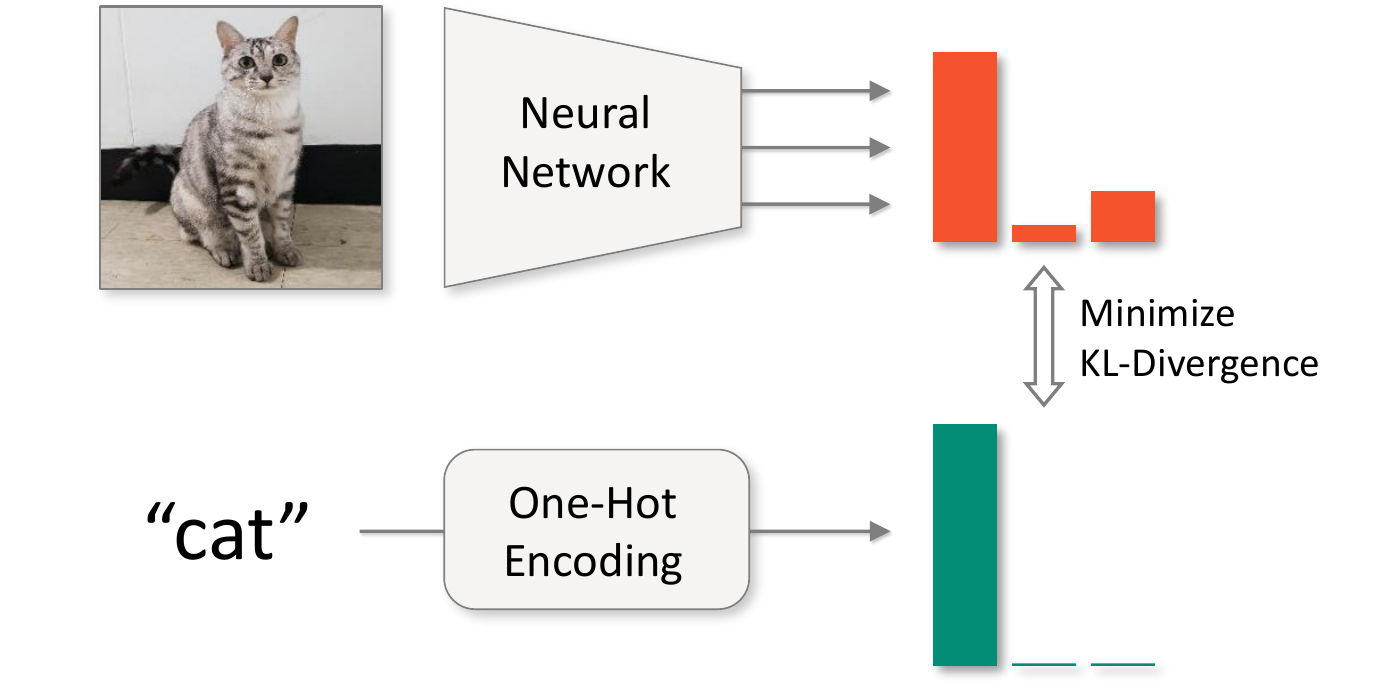}}
    \end{subfigure}
    \hfill
    \begin{subfigure} [Belief matching framework \label{fig1_vbc}]
     {\includegraphics[width=0.40\textwidth]{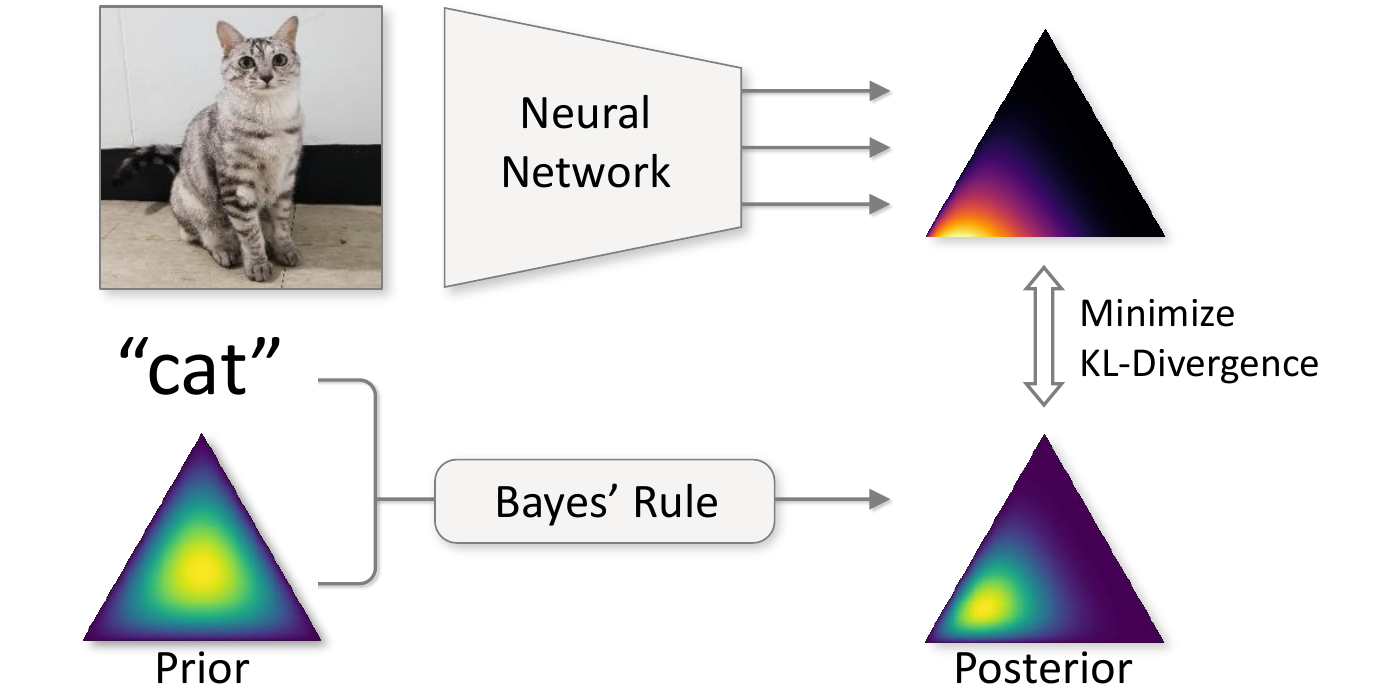}}
    \end{subfigure}
    \hfill
    \caption{Illustration of the difference between softmax cross-entropy loss and belief matching framework when each image is unique in the training set. In softmax cross-entropy loss, the label ``cat" is directly transformed into the target categorical distribution. In belief matching framework, the label ``cat" is combined with the prior Dirichlet distribution over the categorical probability. Then, the Bayes' rule updates the \textit{belief} about categorical probability, which produces the target distribution.
    }
    \label{fig1}
\end{figure*}

\section{Preliminary}
This paper focuses on classification problems in which, given i.i.d. training samples $\mathcal{D} = \left\lbrace \vx^{(i)}, y^{(i)}  \right\rbrace_{i=1}^N \in (\mathcal{X} \times \mathcal{Y})^N$, we construct a classifier $\mathcal{F}: \mathcal{X} \rightarrow \mathcal{Y}$. Here, $\mathcal{X}$ is an input space and $\mathcal{Y} = \left\lbrace1, \cdots, K \right\rbrace$ is a set of labels. 
We denote $\rvx$ and $\ry$ as random variables whose unknown probability distributions generate inputs and labels, respectively. 
Also, we let $\tilde{\rvy}$ be a one-hot representation of $\ry$. 

Let $f^{\mW} : \mathcal{X} \rightarrow \mathcal{X}^\prime$ be a NN with parameters $\mW$ where $\mathcal{X}^\prime = \mathbb{R}^K$ is a logit space. In this paper, we assume $\argmax_j f_j^{\mW} \subseteq \mathcal{F}$ is the classification model where $f_j^{\mW}$ denotes the $j$-th output basis of $f^{\mW}$, and we concentrate on the problem of learning $\mW$. 
Given $((\vx, y) \in \mathcal{D}, f^{\mW})$, a standard minimization loss function is the softmax cross-entropy loss, which applies the softmax to logit and then computes the cross-entropy between a one-hot encoded label and a softmax output (Figure~\ref{fig1_sm}). 
Specifically, the softmax, denoted by $\phi: \mathcal{X}^\prime \rightarrow \triangle^{K-1}$, transforms a logit $f^{\mW}(\vx)$ into a normalized exponential form:
\begin{equation}
    \phi_k(f^{\mW} (\vx)) = \frac{\exp(f^{\mW}_k (\vx))}{\sum_j \exp (f^{\mW}_j (\vx))}
\end{equation}
, and then the cross-entropy loss can be computed by $l_{CE}(\tilde{\vy}, \phi(f^{\mW}(\vx))) = - \sum_k \tilde{\vy}_k \log \phi_k(f^{\mW} (\vx))$.
Here, note that the softmax output can be viewed as a parameter of the categorical distribution, which can be denoted by $\mathcal{P}^C(\phi (f^{\mW}(\vx)))$.

We can formulate the minimization of the softmax cross-entropy loss over $\mathcal{D}$ into a collection of distribution matching problems. 
To this end, let $c^{\mathcal{D}}(\vx)$ be a vector-valued function that counts label frequency at $\vx \in \mathcal{X}$ in $\mathcal{D}$, which is defined as:
\begin{equation}
    c^{\mathcal{D}}(\vx) = \sum_{(\vx^\prime, y^\prime) \in \mathcal{D}} \tilde{\vy}^\prime \mathds{1}_{\{ \vx \}}(\vx^\prime)
\end{equation}
where $\mathds{1}_{\mathcal{A}}(x)$ is an indicator function that takes 1 when $x \in \mathcal{A}$ and 0 otherwise. 
Then, the empirical risk on $\mathcal{D}$ can be expressed as follows:
\begin{multline}
    \hat{\mathcal{L}}_{\mathcal{D}} (\mW) 
    = - \frac{1}{N} \sum_{i=1}^{N} \log \phi_{y^{(i)}}(f^{\mW} (\vx^{(i)})) 
    \\
    = \sum_{\vx \in G( \mathcal{D} )} \frac{\sum_i c_i^{\mathcal{D}}(\vx)}{N} l_{\vx}^{\mathcal{D}}(\mW) + C
\end{multline}
where $G(\mathcal{D})$ is a set of unique values in $\mathcal{D}$, e.g., $G(\{ 1,2,2 \}) = \{ 1,2 \}$, and $C$ is a constant with respect to $\mW$; $l^{\mathcal{D}}_{\vx}(\mW)$ measures the KL divergence between the empirical target distribution and the categorical distribution modeled by the NN at location $\vx$, which is given by:
\begin{equation}
    l^{\mathcal{D}}_{\vx}(\mW) = KL \left( 
        \mathcal{P}^C \left( \frac{c^{\mathcal{D}}(\vx)}{\sum_i c_i^{\mathcal{D}}(\vx)} \right) \parallel \mathcal{P}^C\left( \phi(f^{\mW} (\vx)) \right)
    \right) 
\end{equation}
Therefore, the normalized value of $c^{\mathcal{D}}(\vx)$ becomes the \textit{estimator of a categorical probability} of the target distribution at location $\vx$.
However, directly approximating this target distribution can be problematic.
This is because the estimator uses single or very few samples since most of the inputs are unique or very rare in the training set.

One simple heuristic to handle this problem is label smoothing \citep{szegedy2016rethinking} that constructs a regularized target estimator, in which a one-hot encoded label $\tilde{\vy}$ is relaxed by $(1 - \lambda) \tilde{\vy} + \frac{\lambda}{K}\textbf{1}$ with hyperparameter $\lambda$. 
Under the smoothing operation, the target estimator is regularized by a mixture of the empirical counts and the parameter of the discrete uniform distribution $\mathcal{P}^U$ such that 
$ (1- \lambda) \mathcal{P}^C \left( \frac{c^{\mathcal{D}}(\vx)}{\sum_i c_i^{\mathcal{D}}(\vx)} \right) + \lambda \mathcal{P}^U $.
One concern is that the mixing coefficient is constant with respect to the number of observations, which can possibly prevent the exploitation of the empirical counting information when it is needed.

Another more principled approach is BNNs, which prevents full exploitation of the noisy estimation by balancing the distance to the target distribution with model complexity and maintaining the weight ensemble instead of choosing a single best configuration. 
Specifically, in BNNs with the Gaussian weight prior $\mathcal{N}(\textbf{0}, \tau^{-1} \mI)$, the \textit{score} of configuration $\mW$ is measured by the posterior density $p_{\rmW}(\mW|\mathcal{D}) \propto p(\mathcal{D}| \mW) p_{\rmW}(\mW)$ where we have $\log  p_{\rmW}(\mW) \propto - \tau \parallel \mW \parallel_2^2$. 
Therefore, the complexity penalty term induced by the prior prevents the softmax output from exactly matching a one-hot encoded target.
In modern deep NNs, however, $\parallel \mW \parallel_2^2$ may be poor proxy for the model complexity due to extreme non-linear relationship between weights and outputs \citep{hafner2018reliable,sun2019functional} as well as weight-scaling invariant property of batch normalization \citep{ioffe2015batch}.
This issue may result in poorly regularized predictions, i.e., cannot prevent NNs from the full exploitation of the information contained in the noisy target estimator.

\section{Method} \label{sec_bbm}

\subsection{Constructing Target Distribution}
We propose a Bayesian approach to construct the target distribution for classification, called a belief matching framework (BM; Figure~\ref{fig1_vbc}), in which the categorical probability about a label is regarded as a random variable $\rvz$. 
Specifically, we express the likelihood of $\rvz$ (given $\rvx$) about the label $\ry$ as a categorical distribution $p_{\ry|\rvx, \rvz} = \mathcal{P}^C(\rvz|\rvx)$\footnote{In this paper, $p_{\rvx} = \mathcal{P}(\theta)$ is read as a random variable $\rvx$ follows a probability distribution $\mathcal{P}$ with parameter $\theta$.}. 
Then, specification of the prior distribution over $\rvz | \rvx$ automatically determines the target distribution by means of the Bayesian inference: $p_{\rvz|\rvx,\ry}(\vz) \propto p_{\ry | \rvz, \rvx}(y) p_{\rvz|\rvx}(\vz) $.

We consider a conjugate prior for simplicity, i.e., the Dirichlet distribution. A random variable $\rvz$ (given $\rvx$) following the Dirichlet distribution with concentration parameter vector $\beta$, denoted by $\mathcal{P}^D(\beta)$, has the following density:
\begin{equation}
    p_{\rvz|\rvx}(\vz) = \frac{\Gamma(\beta_0)}{\prod_j \Gamma(\beta_j)} \prod_{k=1}^{K} z_{k}^{\beta_k - 1}
\end{equation}
where $\Gamma(\cdot)$ is the gamma function, $\sum_i z_i = 1$ meaning that $\vz$ belongs to the $K-1$ simplex $\triangle^{K-1}$, $\beta_i > 0, \: \forall i$, and $\beta_0 = \sum_i \beta_i$. 
Here, we have that the mean of $\rvz|\rvx$  is $\beta / \beta_0$ and $\beta_0$ controls the sharpness of the density such that more mass centered around the mean as $\beta_0$ becomes larger.

By the characteristics of the conjugate family, we have the following posterior distribution given $\mathcal{D}$:
\begin{equation}\label{eq:target_dist}
    p_{\rvz | \rvx, \ry} = \mathcal{P}^D (\beta + c^{\mathcal{D}}(\rvx))
\end{equation}
where the target posterior mean is explicitly \textit{smoothed} by the prior belief, and the smoothing operation is performed by the principled way of applying Bayes' rule. 
Specifically, the posterior mean is given by $\frac{1}{\beta_0 + \sum_i c_i^{\mathcal{D}}(\rvx)} (\beta + c^{\mathcal{D}}(\rvx)) $, in which the prior distribution acts as adding pseudo counts. 
We note that the relative strength between the prior belief and the empirical count information becomes adaptive with respect to each data point.

\subsection{Representing Approximate Distribution} \label{sec:approx_dist}
Now, we specify the approximate posterior distribution modeled by the NNs, which aims to approximate $p_{\rvz|\rvx, \ry}$. 
In this paper, we model the approximate posterior as the Dirichlet distribution.
To this end, we use an exponential function $g (x) = \exp(x)$ to transform logits to the concentration parameter of $\mathcal{P}^{D}$, and we let $\alpha^{\mW} = \exp \circ f^{\mW}$. 
Then, the NN represents the density over $\triangle^{K-1}$ as follows:
\begin{equation}\label{eq:dirichlet_density}
    q_{\rvz|\rvx}^{\mW}(\vz) = 
    \frac{\Gamma(\alpha^{\mW}_0(\rvx))}
        {\prod_j \Gamma( \alpha^{\mW}_j(\rvx))} 
    \prod_{k=1}^{K} z_{k}^{\alpha^{\mW}_k(\rvx) - 1}
\end{equation}
where $\alpha_0^{\mW}(\rvx) = \sum_i \alpha^{\mW}_i(\rvx)$.

From \eqref{eq:dirichlet_density}, we can see that outputs under BM encode much more information compared to those under the softmax. 
Specifically, it can be easily shown that the approximate posterior mean corresponds to the softmax. 
In this regard, BM enables neural networks to represent more rich information in their outputs, i.e., the density over $\triangle^{K-1}$ itself not just a single point on it such as the mean.
This capability allows capturing more diverse characteristics of predictions at different locations, such as how much concentrate its density around the center mass point, which can be extremely helpful in many applications.
For instance, BM gives a more sophisticated measure of the difference between predictions of two neural networks, which can benefit the consistency-based loss for semi-supervised learning as we will show in section~\ref{sec:semi_supervised_learning}.
Besides, BM represents a more sophisticated measure of predictive uncertainty based on the density over simplex, such as mutual information.


From the perspective of learning the target distribution, BM can be considered as a generalization of softmax in terms of changing the moment matching problem to the distribution matching problem in $\mathcal{P}(\triangle^{K-1})$.
To understand the distribution matching objective in BM, we reformulate \eqref{eq:dirichlet_density} as follows:
\begin{multline}
    q_{\rvz|\rvx}^{\mW}(\vz)
    \propto \exp \left(
        \sum_k \alpha^{\mW}_k (\rvx) \log z_{k} - \sum_k  \log z_{k}
    \right) \\
    \propto \exp \left(
        - l_{CE}(\phi(f^{\mW}(\rvx)), \vz) 
            + \frac{KL(\mathcal{P}^U \parallel \mathcal{P}^C(\vz))}{\alpha^{\mW}_0 (\rvx) / K}  
        \right)
\end{multline}
In the limit of $q_{\rvz|\rvx}^{\mW} \rightarrow p_{\rvz | \rvx, \ry}$, mean of the target posterior (\eqref{eq:target_dist}) becomes a virtual label, for which individual $\vz$ ought to match;
the penalty for ambiguous configuration $\vz$ is determined by the number of observations.
Therefore, the distribution matching in BM can be thought of as \textit{learning to score a categorical probability} based on closeness to the target posterior mean, in which exploitation of the closeness information is automatically controlled by the data.

\subsection{Distribution Matching}
We have defined the target distribution $p_{\rvz | \rvx, \ry }$ and the approximate distribution modeled by the neural network $q_{\rvz | \rvx}^{\mW}$.  
We now present a solution to the distribution matching problem with maximizing the evidence lower bound (ELBO), defined by $l_{EB}(\ry, \alpha^{\mW}(\vx)) = \mathbb{E}_{q_{\rvz|\rvx}^{\mW} } [ \log p(\ry | \rvx, \rvz)] - KL(q^{\mW}_{\rvz | \rvx} \parallel p_{\rvz | \rvx} )$.
Using the ELBO can be motivated by the following equality \citep{jordan1999introduction}:
\begin{multline}
    \log p(\ry | \rvx) 
    = \int q_{\rvz|\rvx}^{\mW}(\vz) \log \left( 
        \frac{p(\ry, \vz| \rvx)}{p (\vz | \rvx, \ry)}
    \right) d\vz \\
    = l_{EB}(\ry, \alpha^{\mW}(\vx)) - KL (q_{\rvz|\rvx}^{\mW} \parallel  p_{\rvz|\rvx, \ry}) 
\end{multline}
where we can see that maximizing $l_{EB}(\ry, \alpha^{\mW}(\rvx))$ corresponds to minimizing $ KL (q_{\rvz|\rvx}^{\mW} \parallel  p_{\rvz|\rvx, \ry})$, i.e., \textit{matching the approximate distribution to the target distribution}, because the KL-divergence is non-negative and $ \log p(\ry | \rvx)$ is a constant with respect to $\mW$. 
Here, each term in the ELBO can be analytically computed by:
\begin{multline}\label{eq:loss_evi}
    \mathbb{E}_{q_{\rvz|\rvx} } \left[ 
        \log p(\ry | \rvx, \rvz)
    \right] \\
    = \mathbb{E}_{q_{\rvz|\rvx} } [\log \rz_{\ry}]
    = \psi (\alpha^{\mW}_{\ry} (\rvx)) - \psi (\alpha^{\mW}_0(\rvx))
\end{multline}
where $\psi(\cdot)$ is the digamma function (the logarithmic derivative of $\Gamma(\cdot)$), and 
\begin{multline}\label{eq:loss_kl}
    KL(q^{\mW}_{\rvz | \rvx} \parallel p_{\rvz | \rvx} ) 
    = \log 
    \frac{\Gamma (\alpha^{\mW}_0 (\rvx)) \prod_k \Gamma (\beta_k)}
    {\prod_k \Gamma (\alpha^{\mW}_k (\rvx)) \Gamma (\beta_0)}  \\ 
    + \sum_k  \left( \alpha^{\mW}_k (\rvx)- \beta_k \right) \left( \psi (\alpha^{\mW}_k (\rvx)) - \psi (\alpha^{\mW}_0 (\rvx)) \right)
\end{multline}
where $p_{\rvz|\rvx}$ is assumed to be an input independent conjugate prior for simplicity; that is, $p_{\rvz | \rvx} = \mathcal{P}^D(\beta)$.
With this analytical solution, we maximizes the ELBO with mini-batch approximation, which gives the following loss function:  $\mathcal{L}(\mW) = \mathbb{E}_{\rvx, \ry}[ l_{EB}(\ry, \alpha^{\mW}(\rvx))] \approx \frac{1}{m} \sum_{i=1}^{m} l_{EB}(y^{(i)}, \alpha^{\mW}(\vx^{(i)}))$.
We note that computations of the ELBO and its gradient have a complexity of $\mathcal{O}(K)$ per sample, which is equal to those of softmax.
This means that BM can preserve the scalability and the efficiency of the existing baseline. 
We also note that the analytical solution of the ELBO under BM allows to implement the distribution matching loss as \textit{a plug-and-play loss function} applied to the logit directly.

\subsection{On Prior Distributions} \label{on_prior}
The success of the Bayesian approach largely depends on how we specify the prior distribution due to its impact on the resulting posterior distribution.
For example, the target posterior mean in \eqref{eq:target_dist} becomes the counting estimator as $\beta_0 \rightarrow 0$. 
On the contrary, as $\beta_0$ becomes higher, the effect of empirical counting information is weakened, and eventually disappeared in the limit of $\beta_0 \rightarrow \infty$. 
Therefore, considering that most of the inputs are unique in $\mathcal{D}$, choosing small $\beta_0$ is appropriate for prevents the resulting posterior distribution from being dominated by the prior\footnote{In an ideal fully Bayesian treatment, $\beta$ can be modeled hierarchically, and we left this as future research.}. 

However, a prior distribution with small $\beta_0$ implicitly makes $\alpha_0^{\mW}( \rvx)$ small, which poses significant challenges on the gradient-based optimization. 
This is because the gradient of the ELBO is notoriously large in the small-value regimes of $\alpha_0^{\mW}(\rvx)$, e.g., $\psi^\prime(0.01) > 10000$.
In addition, our various building blocks including normalization \citep{ioffe2015batch}, initialization \citep{he2015delving}, and architecture \citep{he2016deep} are implicitly or explicitly designed to make $\mathbb{E}[f^{\mW}(\rvx)] \approx \textbf{0}$; that is, $\mathbb{E}[\alpha^{\mW}(\rvx)] \approx \textbf{ 1}$. Therefore, making $\alpha_0^{\mW}(\rvx)$ small can be wasteful or requires huge modifications to the existing building blocks.
Also, $\mathbb{E}[\alpha^{\mW}(\rvx)] \approx \textbf{1}$ is encouraged in a sense of natural gradient \citep{amari1998natural}, which improves the conditioning of Fisher information matrix \citep{schraudolph1998accelerated,lecun1998efficient, raiko2012deep, wiesler2014mean}.


In order to resolve the gradient-based optimization challenge in learning the posterior distribution while preventing dominance of the prior distribution, we set $\beta = \textbf{1}$ for the prior distribution and then multiply $\lambda$ to the KL divergence term in the ELBO: $l^{\lambda}_{EB}(\ry, \alpha^{\mW}(\rvx)) =  \mathbb{E}_{q_{\rvz|\rvx} } \left[ \log p(\ry | \rvx, \rvz) \right] - \lambda KL(q^{\mW}_{\rvz | \rvx} \parallel \mathcal{P}^D(\textbf{1}) )$.  
This trick significantly stabilizes the optimization process, while making a local optimal point remains unchanged.
To see this, we can compare the gradients of the ELBO and the lambda multiplied ELBO:
\begin{multline}\label{eq:grad_logit}
    \frac{\partial l_{EB}(\ry, \alpha^{\mW}(\rvx))}{\partial \alpha_k^{\mW} (\vx)}
    = \left(\tilde{\rvy}_k - ( \alpha_k^{\mW}(\vx) - \beta_k)\right) \psi^\prime (\alpha_k^{\mW}(\vx)) \\
    - \left(1 - (\alpha_0^{\mW}(\vx) - \beta_0 )\right) \psi^\prime (\alpha_0^{\mW}(\vx)) 
\end{multline}
\begin{multline}\label{eq:new_grad_logit}
    \frac{\partial l^{\lambda}_{EB}(\ry, \alpha^{\mW}(\rvx))}{\partial \alpha_k^{\mW}(\rvx)}  
    = \left(\tilde{\rvy}_k - ( \tilde{\alpha}_k^{\mW} (\rvx)- \lambda)\right) 
        \frac{\psi^\prime (\tilde{\alpha}_k^{\mW}(\rvx))}{\psi^\prime (\tilde{\alpha}^{\mW}_0 (\rvx))} \\
    - \left(1 - (\tilde{\alpha}_0^{\mW}(\rvx) - \lambda K )\right) 
\end{multline}
where $\tilde{\alpha}_k^{\mW} (\rvx) = \lambda \alpha_k^{\mW}(\rvx) $. 
Here, we can see that a local optimal in \eqref{eq:grad_logit} is achieved when $\alpha^{\mW}(\rvx) = \beta + \tilde{\rvy}$ and a local optima for \eqref{eq:new_grad_logit} is $\alpha^{\mW}(\rvx)= 1 + \frac{1}{\lambda} \tilde{\rvy}$.
Therefore, a ratio between $\alpha_i^{\mW}(\rvx)$ and $\alpha_j^{\mW}(\rvx)$ equal to those of a local optimal point in \eqref{eq:grad_logit} for every pair of $i$ and $j$.
In this regard, searching for $\lambda$ with $l^{\lambda}_{EB}(\ry, \alpha^{\mW}(\rvx))$ and then multiplying $\lambda$ after training corresponds to the process of searching for the prior distribution's parameter $\beta$ with $\mathcal{L}(\mW)$.

\section{Related Work}
BNNs are the dominant approach for applying Bayesian principles in neural networks. 
Because BNNs require the intractable posterior inference, many posterior approximation schemes have been developed to reduce the approximation gap and improve scalability (e.g., VI \citep{graves2011practical, blundell2015weight, wu2019deterministic} and stochastic gradient Markov Chain Monte Carlo \citep{welling2011bayesian,ma2015complete,gong2018metalearning}). 
However, even with these novel approximation techniques, BNNs are not scalable to state-of-the-art architectures in large-scale datasets or they often reduce the generalization performance in practice, which impedes the wide adoption of BNNs despite their numerous potential benefits.

Other approaches avoid explicit modeling of the weight posterior distribution.
MC dropout \citep{gal2016dropout} reinterprets the dropout \citep{srivastava2014dropout} as an approximate VI, which retains the standard NN training procedure and modifies only the inference procedure for posterior MC approximation.
In a similar spirit, some approaches \citep{mandt2017stochastic,zhang2018noisy,maddox2019simple,osawa2019practical} sequentially estimate the mean and covariance of the weight posterior distribution by using gradients computed at each step.
As different from the BNNs, Deep kernel learning \citep{wilson2016kernel,wilson2016stochastic} places Gaussian processes (GPs) on top of the ``deterministic" NNs, which combines NNs' capability of handling complex high dimensional data and GPs' capability of principled uncertainty representation and robust extrapolation.

Non-Bayesian approaches also help to resolve the limitations of softmax. 
\citet{lakshminarayanan2017simple} propose an ensemble-based method to achieve better uncertainty representation and improved self-calibration. 
Both \citet{guo2017calibration} and \citet{neumann2018relaxed} proposed temperature scaling-based methods for post-hoc modifications of softmax for improved calibration. 
To improve generalization by penalizing over-confidence, \citet{pereyra2017regularizing} propose an auxiliary loss function that penalizes low predictive entropy, and \citet{szegedy2016rethinking} and \citet{xie2016disturblabel} consider the types of noise included in ground-truth labels. 

We also note that some recent studies use NNs to model the concentration parameter of the Dirichlet distribution but with a different purpose than BM. 
\citet{sensoy2018evidential} uses the loss function of explicitly minimizing prediction variances on training samples, which can help to produce high uncertainty prediction for out-of-distribution (OOD) or adversarial samples.
Prior network \citep{malinin2018predictive} investigates two types of auxiliary losses computed on in-distribution and OOD samples, respectively. 
Similar to prior network, \citet{chen2018variational} considers an auxiliary loss computed on adversarially generated samples.

\section{Experiment}
In this section, we show versatility of BM through extensive empirical evaluations. 
We first verify its improvement of the generalization error in image classification tasks (section~\ref{benchmark}).
We then verify whether BM inherits the advantages of the Bayesian method by placing the prior distribution \textit{only} on the label categorical probability (section~\ref{sec:uncertainty_representation}).
We conclude this section by providing further applications that shows versatility of BM. To support reproducibility, we release our code at: \url{https://github.com/tjoo512/belief-matching-framework}.
We performed all experiments on a single workstation with 8 GPUs (NVIDIA GeForce RTX 2080 Ti).

Throughout all experiments, we employ various large-scale models based on a residual connection \citep{he2016deep}, which are the standard benchmark models in practice.
For fair comparison and reducing burden of hyperparameter search, we fix experimental configurations to the reference implementation of corresponding architecture. 
However, we additionally use an initial learning rate warm-up and gradient clipping, which are extremely helpful for stable training of BM. 
Specifically, we use learning rates of [0.1$\epsilon$, 0.2$\epsilon$, 0.4$\epsilon$, 0.6$\epsilon$, 0.8$\epsilon$] for first five epochs when the reference learning rate is $\epsilon$ and clip gradient when its norm exceeds 1.0. 
Without these methods, we had difficulty in training deep models, e.g., ResNet-50, due to gradient explosion at an initial stage of training.


We compare BM to following baseline methods: softmax, which is our primary object to improve; MC dropout with 100 MC samples, which is a simple and efficient BNN; deep ensemble with five NNs, which greatly improves the uncertainty representation ability. 
While there are other methods using NNs to model the Dirichlet distribution \citep{sensoy2018evidential,malinin2018predictive,malinin2019reverse}, we note that these methods are not scalable to ResNet. 
Similarly, we observe that training a mixture of Dirichlet distributions \citep{wu2019quantifying} with ResNet is subject to the gradient explosion, even with a 10x lower learning rate. 
Besides, BNNs with VI (or MCMC) are not directly comparable to our approach due to their huge modifications to existing baselines. 
For example, \citet{heek2019bayesian} replace batch normalization and ReLU, use additional techniques (2x more filters, cyclic learning rate, snapshot ensemble), and require almost 10x more computations on ImageNet to converge.

\begin{table}[t]
\caption{Test classification error rates on CIFAR. 
Here, we split a train set of 50K examples into a train set of 40K examples and a validation set of 10K example.
Numbers indicate $\mu \pm \sigma$ computed across five trials, and boldface indicates the minimum mean error rate. Model and hyperparameter are selected based on validation error rates.
We searched for the coefficients of BM over $\left\lbrace 0.01, 0.003, 0.001 \right\rbrace$ and MC dropout over $\left\lbrace 0.1, 0.2, 0.5 \right\rbrace$. 
}
\vskip 0.1in
\begin{center}
\begin{small}
\begin{sc}\label{cifar}
\begin{tabular}{llll}
\toprule
Model & Method & C-10 & C-100 \\
\midrule
Res-18 & Softmax & 6.13 $\pm 0.13$  & 26.44 $\pm 0.33 $ \\
& MC Drop (last)        & 6.13 $\pm 0.08$  & 26.15 $\pm 0.10$  \\
& MC Drop (all)        & 6.50 $\pm 0.14$  & 27.32 $\pm 0.45$  \\
& BM                & \textbf{5.93} $\pm 0.07$ & \textbf{24.19} $\pm 0.34 $ \\
\midrule
Res-50 & Softmax & 5.76 $\pm 0.06$ & 25.00 $\pm 0.23 $ \\
& MC Drop (last)        & 5.75 $\pm 0.22$   & 25.17 $\pm 0.09$  \\
& MC Drop (all)        & 5.84 $\pm 0.23 $   & 26.74 $\pm 0.37$  \\
& BM                & \textbf{5.59} $\pm 0.05 $ & \textbf{23.86} $\pm 0.37 $ \\
\bottomrule
\end{tabular}
\end{sc}
\end{small}
\end{center}
\vskip -0.1in
\end{table}

\subsection{Generalization Performance} \label{benchmark}
We evaluate the generalization performance of BM on CIFAR \citep{cifar} with the pre-activation ResNet \citep{he2016identity}. 
CIFAR-10 and CIFAR-100 contain 50K training and 10K test images, and each 32x32x3-sized image belongs to one of 10 categories in CIFAR-10 and one of 100 categories in CIFAR-100. 
Table~\ref{cifar} lists the classification error rates of the softmax cross-entropy loss, BM, and MC-dropout. 
In all configurations, BM consistently achieves the best generalization performance on both datasets. 
On the other hand, last-layer MC dropout sometimes results in higher generalization errors than softmax and all-layer MC-dropout significantly increases error rates even though they consume 100x more computations for inference.

We next perform a large-scale experiment using ResNext-50 32x4d and ResNext-101 32x8d \citep{xie2017aggregated} on ImageNet \citep{ILSVRC15}. 
ImageNet contains approximately 1.3M training samples and 50K validation samples, and each sample is resized to 224x224x3 and belongs to one of the 1K categories;
that is, the ImageNet has more categories, a larger image size, and more training samples compared to CIFAR, which may enable a more precise evaluation of methods.
Consistent with the results on CIFAR, BM improves test errors of softmax (Table~\ref{imgnet}).
This result is appealing because improving the generalization error of deep NNs on large-scale datasets by adopting a Bayesian principle without computational overhead has rarely been reported in the literature.

\begin{table}[t]
\caption{Classification error rates on the ImageNet. 
Here, we use only $\lambda = 0.001$ for ResNext-50 and $\lambda = 0.0001$ for ResNext-101, and measure the validation error rates directly. 
We report the result obtained by single experiment due to computational constraint. }
\vskip 0.1in
\begin{center}
\begin{small}
\begin{sc}\label{imgnet}
\begin{tabular}{llll}
\toprule
Model & Method & Top1 & Top5 \\
\midrule
ResNext-50 & Softmax & 22.23  & 6.36  \\
& BM                & \textbf{22.03} & \textbf{6.29} \\
\midrule
ResNext-101 & Softmax & 20.72  & 5.59  \\
& BM                & \textbf{20.23} & \textbf{5.26} \\
\bottomrule
\end{tabular}
\end{sc}
\end{small}
\end{center}
\vskip -0.1in
\end{table}

\begin{figure}
    \centering
    \begin{subfigure}
     {\includegraphics[width=0.23\textwidth]{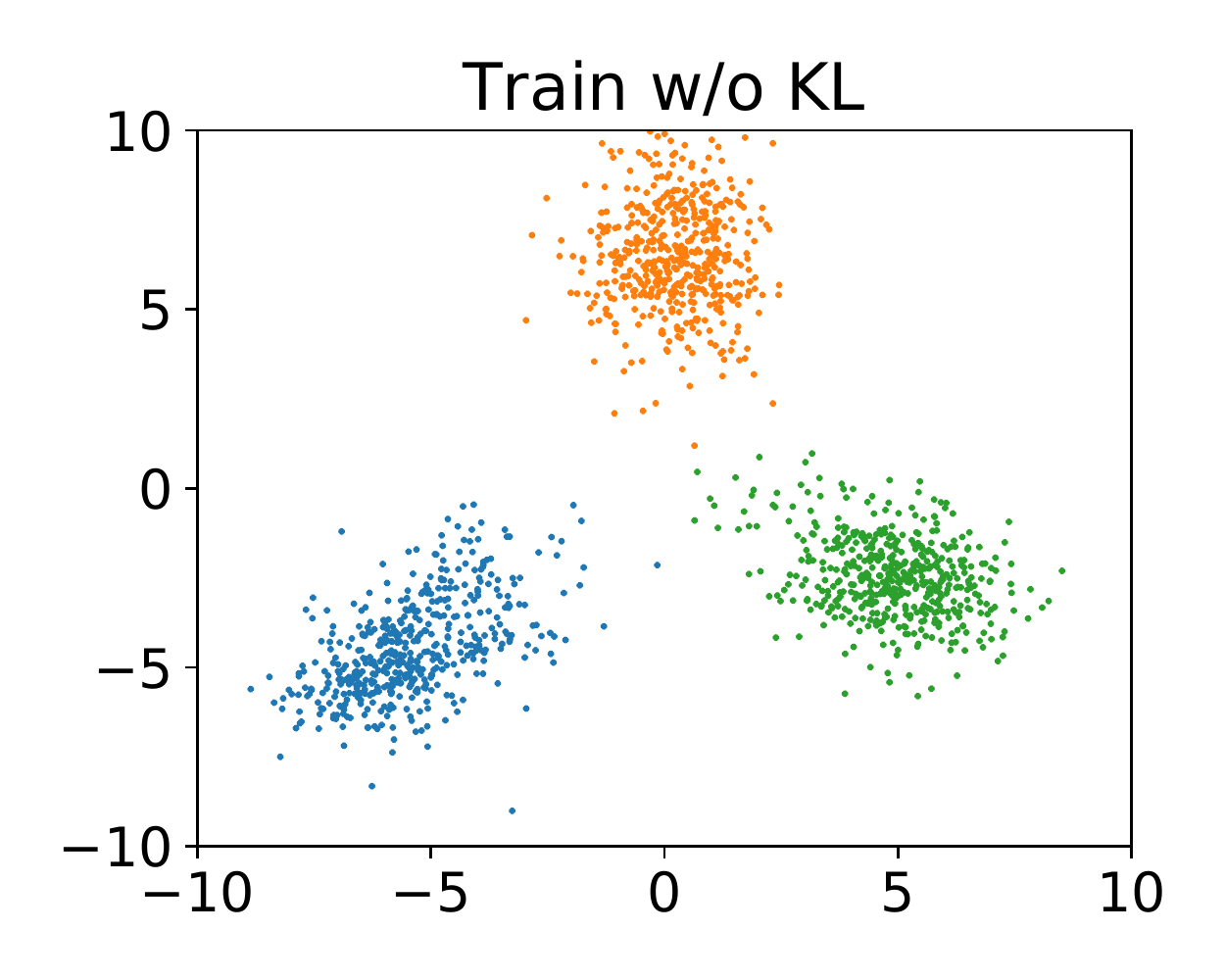}}
    \end{subfigure}
    \hfill
    \begin{subfigure}
     {\includegraphics[width=0.23\textwidth]{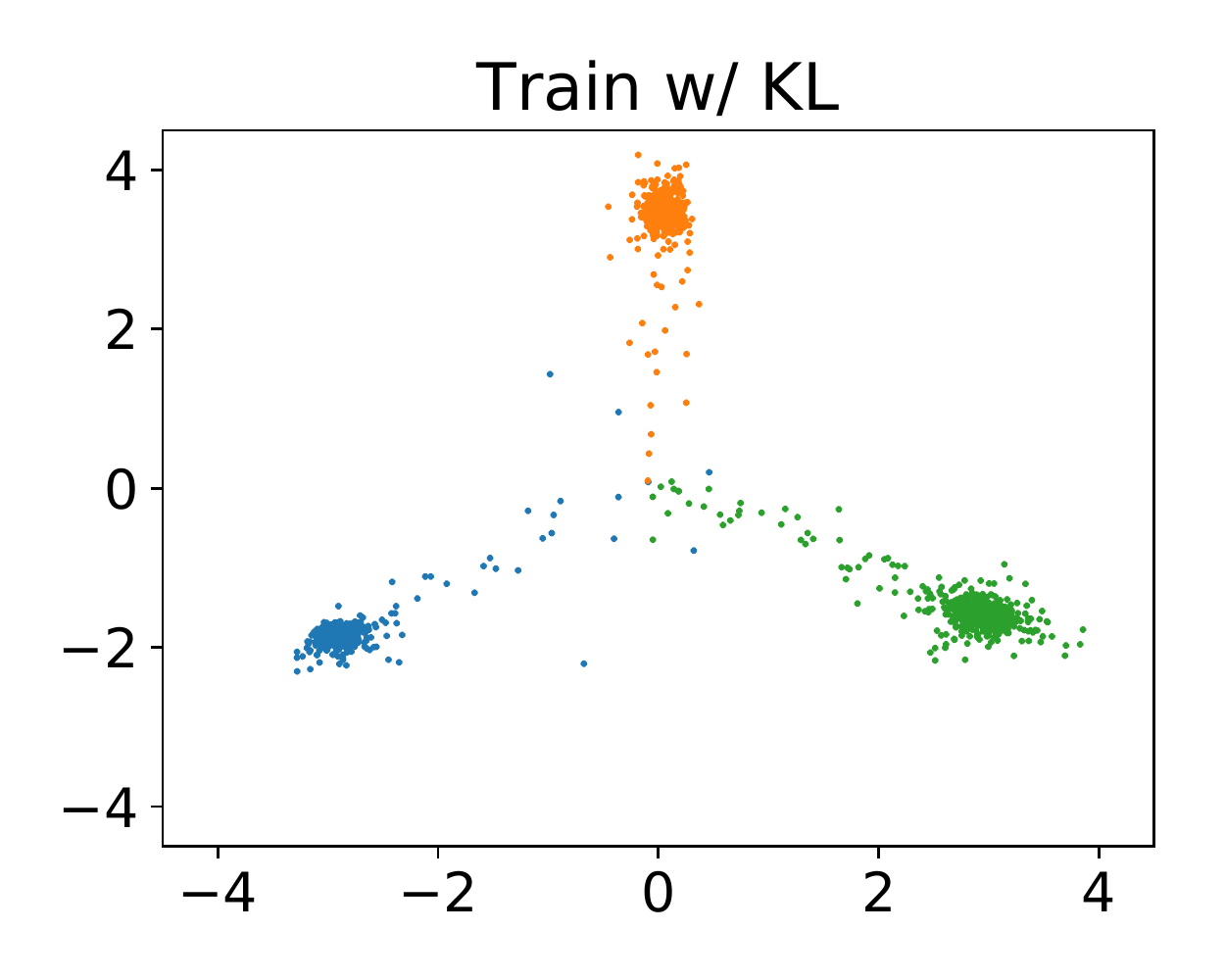}}
    \end{subfigure}
    \hfill
    \caption{
    Penultimate layer's activations of examples belonging to one of three classes (beaver, dolphin, and otter; indexed by 0,1,2 in CIFAR-100).}
    \label{representation}
\end{figure}

\paragraph{Regularization Effect of Prior}
In theory, BM has two regularization effects, which may explain the generalization performance improvements under BM: the prior distribution, which smooths the target posterior mean by adding pseudo counts, and computing the distribution matching loss by averaging of all possible categorical probabilities. 
In this regard, the ablation of the KL term in the ELBO helps to examine these two effects separately, which removes only the effect of the prior distribution. 

We first examine its impact on the generalization performance by training a ResNet-50 on CIFAR without the KL term. 
The resulting test error rates were $\textbf{5.68\%}$ on CIFAR-10 and $\textbf{24.69}\%$ on CIFAR-100.
These significant reductions in generalization performances indicates the powerful regularization effect of the prior distribution (cf. Table~\ref{cifar}). 
The result that BM without the KL term still achieves lower test error rates compared to softmax demonstrates the regularization effect of considering all possible categorical probabilities by the Dirichlet distribution instead of choosing single categorical probability.

Considering the role of the prior distribution on smoothing the posterior mean, we conjecture that the impact of the prior distribution is similar to the effect of label smoothing. 
In \citet{muller2019does}, it is shown that label smoothing makes learned representation reveal tight clusters of data points within the same classes and smaller deviations among the data points.
Inspired by this result, we analyze the activations in the penultimate layer with the visualization method proposed in \citet{muller2019does}. 
Figure~\ref{representation} illustrates that the prior distribution significantly reduces the value ranges of the activations of data points, which implies the implicit function regularization effect considering that the $L^p$ norm of $f^{\mW} \in L^p(\mathcal{X}) $ can be approximated by $\parallel f^{\mW} \parallel_p \approx \left( \frac{1}{N} \sum_i |f^{\mW}(\vx^{(i)})|^p\right)^{1/p} $.
Besides, Figure~\ref{representation} shows that the prior distribution makes activations belong to the same class to form much tighter clusters, which can be thought of as the implicit manifold regularization effect.
To see this, assume that two images belonging to the same class have close distance in the data manifold.
Then, the difference between logits of same class examples becomes a good proxy for the gradient of $f^{\mW}$ along the data manifold since the gradient measures changes in the output space with respect to small changes in the input space. 

\begin{figure}
    \centering
    \includegraphics[width=0.25\textwidth]{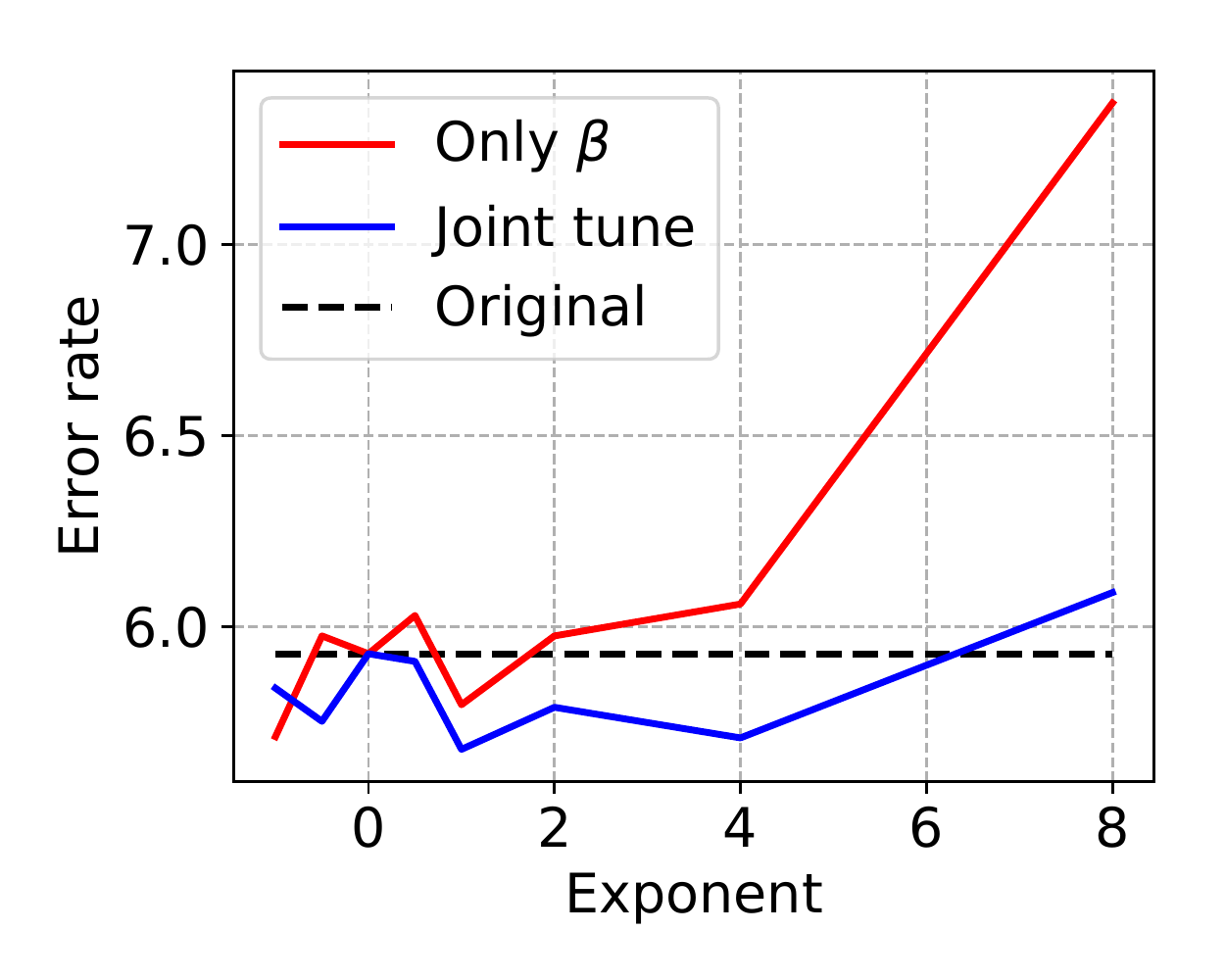}
    \caption{
    Impact of $\beta$ on generalization performance. 
    We exclude the ranges $\beta < \exp(-2)$ and $\beta > \exp(8)$ because the ranges result in gradient explosion under the strategy of changing only $\beta$
    }
    \label{beta_test}
\end{figure}

\paragraph{Impact of $\beta$}
In section~\ref{on_prior}, we claimed that a value of $\beta$ is implicitly related to the distribution of logit values, and its extreme value can be detrimental to the training stability. 
We verify this claim by training ResNet-18 on CIFAR-10 with different values of $\beta$. 
Specifically, we examine two strategies of changing $\beta$: modifying only $\beta$ or jointly modifying the lambda proportional to $1 / \beta$ to match local optima (cf. section~\ref{on_prior}).
As a result, we obtain a robust generalization performance in both strategies when $\beta \in [\exp(-1), \exp(4) ]$ (Figure~\ref{beta_test}). 
However, when $\beta$ becomes extremely small ($\exp(-2)$ when changing only $\beta$ and $\exp(-8)$ when jointly tuning $\lambda$ and $\beta$), the gradient explosion occurs due to extreme slope of the digamma near 0.
Conversely, when we increase only $\beta$ to extremely large value, the error rate increases by a large margin (7.37) at $\beta=\exp(8)$, and eventually explodes at $\beta = \exp(16)$. 
This is because large beta increases the values of activations, so the gradient with respect to parameters explodes. 
Under the joint tuning strategy, such a high values region makes $\lambda \approx 0$, which removes the impact of the prior distribution.

\subsection{Uncertainty Representation} \label{sec:uncertainty_representation}
One of the most attractive benefits of Bayesian methods is their ability to represent the uncertainty about their predictions.
In a naive sense, uncertainty representation ability is the ability to ``know what it doesn't know." 
For instance, models having a good uncertainty representation ability would increase some form of predictive uncertainty on misclassified examples compared to those on correctly classified examples.
This ability is extremely useful in both real-world applications and downstream tasks in machine learning.
For example, underconfident NNs can produce many false alarms, which makes humans ignore the predictions of NNs; conversely, overconfident NNs can exclude humans from the decision-making loop, which results in catastrophic accidents.
Also, better uncertainty representation enables balancing exploitation and exploration in reinforcement learning \citep{gal2016dropout} and detecting OOD samples \citep{malinin2018predictive}.

We evaluate the uncertainty representation ability on both in-distribution and OOD datasets.
Specifically, we measure the calibration performance on in-distribution test samples, which examines a model's ability to match its probabilistic output associated with an event to the actual long-term frequency of the event \citep{dawid1982well}. 
The notion of calibration in NNs is associated with how well its confidence matches the actual accuracy; e.g., we expect the average accuracy of a group of predictions having the confidence around $0.7$ to be close to  $70\%$.
We also examine the predictive uncertainty for OOD samples. 
Since the examples belong to none of the classes seen during training, we expect neural networks to produce outputs of ``I don't know."



\begin{figure}
    \centering
    \begin{subfigure}[CIFAR-10]
     {\includegraphics[width=0.23\textwidth]{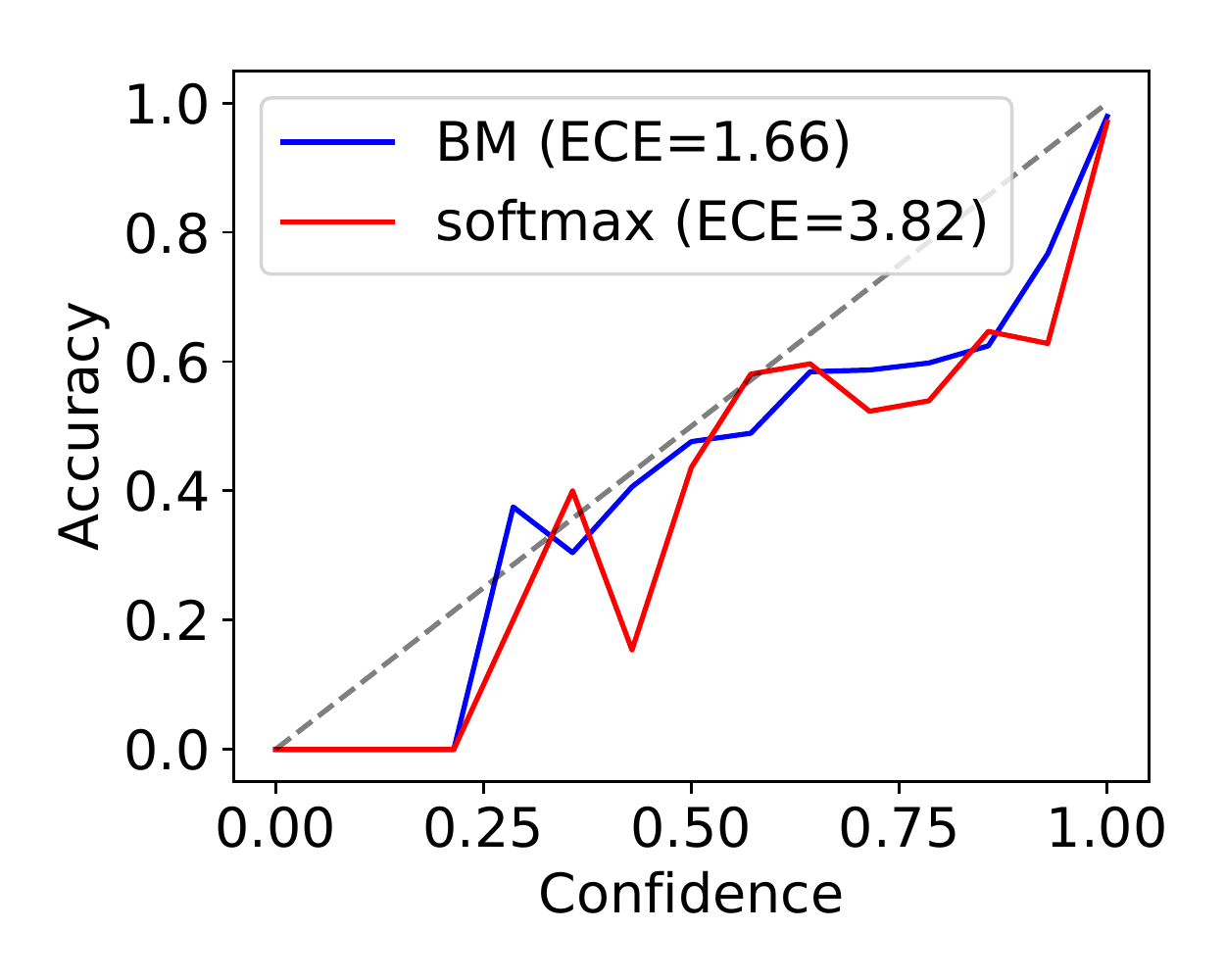}}
    \end{subfigure}
    \hfill
    \begin{subfigure}[CIFAR-100]
     {\includegraphics[width=0.23\textwidth]{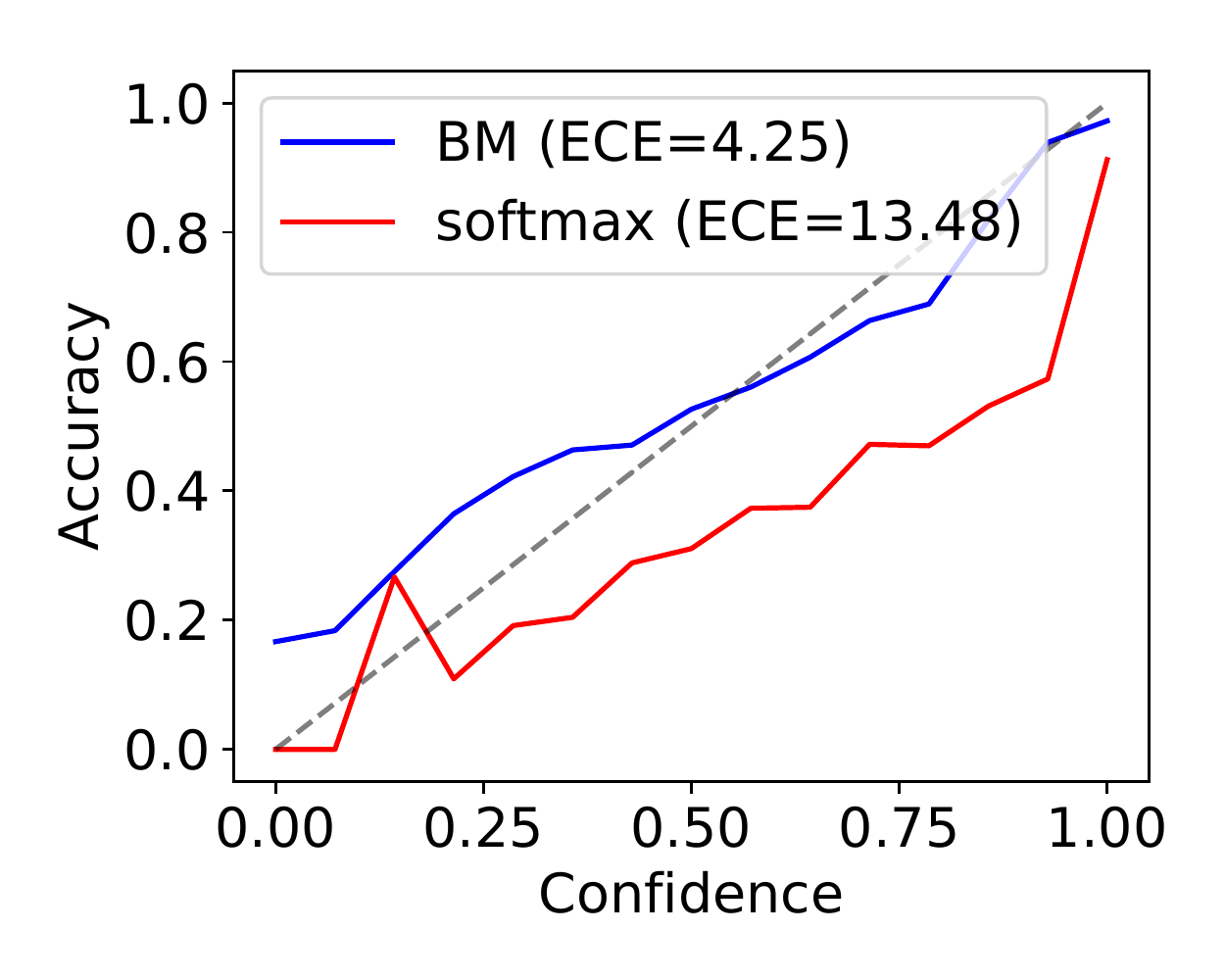}}
    \end{subfigure}
    \hfill
    \caption{Reliability plots of ResNet-50 with BM and softmax. 
    Here, ECE is computed with 15 groups.}
    \label{calibration}
\end{figure}

\paragraph{In-Distribution Uncertainty} 
We measure the calibration performance by the expected calibration error (ECE; \citeauthor{naeini2015obtaining}, \citeyear{naeini2015obtaining}), in which $\max_i \phi_i (f^{\mW}(\vx))$ is regarded as a prediction confidence for the input $\vx$.
ECE is calculated by grouping predictions based on the confidence score and then computing the absolute difference between the average accuracy and average confidence for each group; that is, the ECE of $f^{\mW}$ on $\mathcal{D}$ with $M$ groups is as follows:
\begin{equation}
 ECE^M(f^{\mW}, \mathcal{D}) = \sum_{i=1}^{M} \frac{| \mathcal{G}_i|}{|\mathcal{D}|}| 
\text{acc}(\mathcal{G}_i) 
- \text{conf}(\mathcal{G}_i) | \end{equation}
where $\mathcal{G}_i$ is a set of samples in the $i$-th group, defined as $\mathcal{G}_i = \left\lbrace j :i / M < \max_k \phi_k(f^{\mW}( \vx^{(j)})) \leq (1 + i) / M \right\rbrace$, 
$\text{acc}(\mathcal{G}_i)$ is an average accuracy in the $i$-th group, and $\text{conf}(\mathcal{G}_i)$ is an average confidence in the $i$-th group.


We analyze the calibration property of ResNet-50 examined in section~\ref{benchmark}. 
As Figure~\ref{calibration} presents, BM's predictive probability is well matched to its accuracy compared to softmax--that is, BM improves the calibration property of NNs.
Specifically, BM improves ECE of softmax from 3.82 to 1.66 on CIFAR-10 and from 13.48 to 4.25 CIFAR-100, respectively.
These improvements are comparable to the deep ensemble, which achieves 1.04 on CIFAR-10 and 3.54 on CIFAR-100 with 5x more computations for both training and inference. 
In the case of all-layer MC dropout, ECE decreases to 1.50 on CIFAR-10 and 9.76 on CIFAR-100, however, these improvements require to compromise the generalization performance. 
On the other hand, the last-layer MC dropout, which often improves the generalization performance, does not show meaningful ECE improvements (3.78 on CIFAR-10 and 13.52 on CIFAR-100).
We note that there are also post-hoc solutions improving calibration performance, e.g., temperature scaling \citep{guo2017calibration}. However, these methods often require an additional dataset for tuning the behavior of NNs, which may prevent the exploitation of entire samples to train NNs.

\begin{figure}
    \begin{subfigure}[Softmax \label{sm_uncer}]
     {\includegraphics[width=0.23\textwidth]{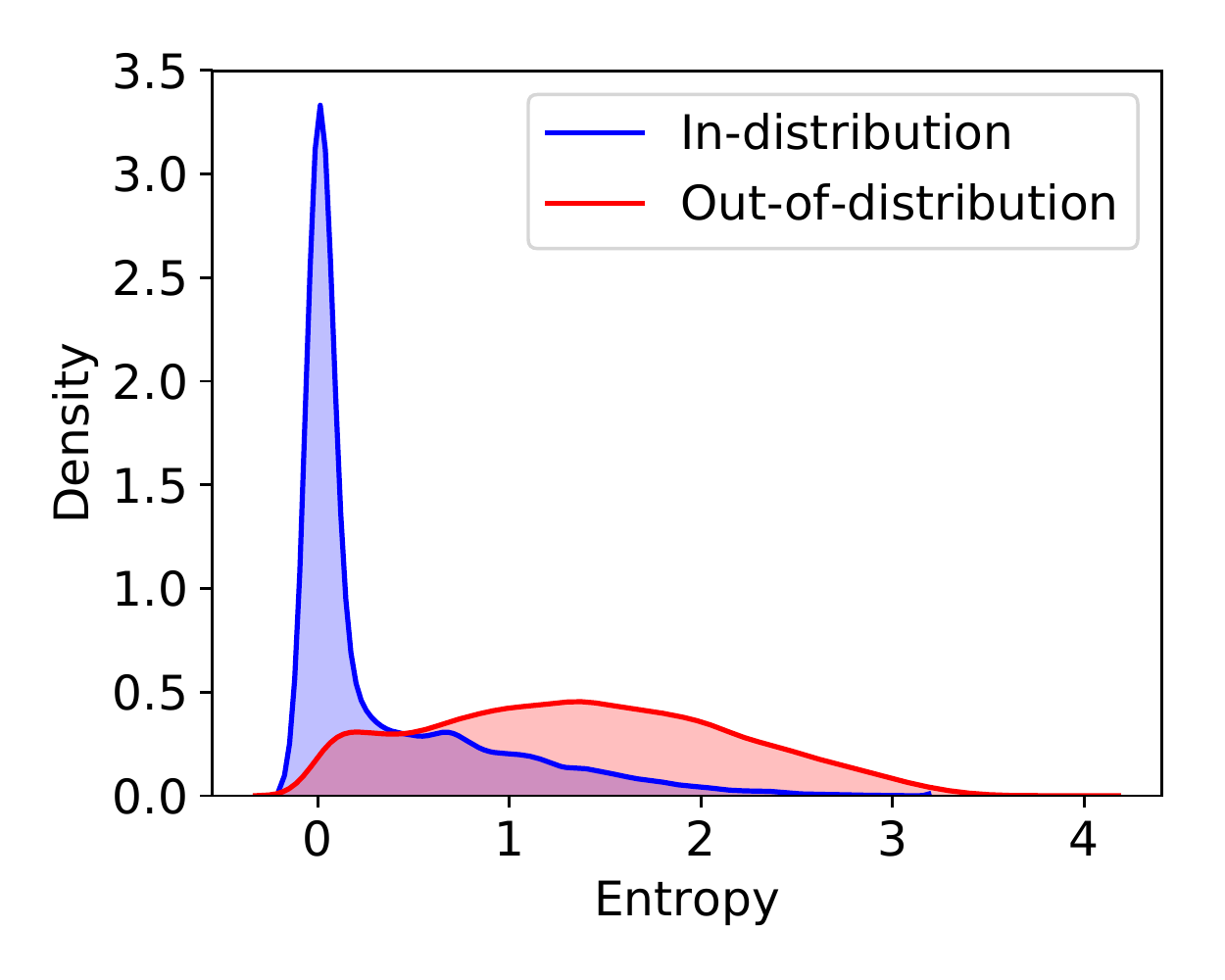}}
    \end{subfigure}
    \begin{subfigure}[BM \label{vc_uncer}]
     {\includegraphics[width=0.23\textwidth]{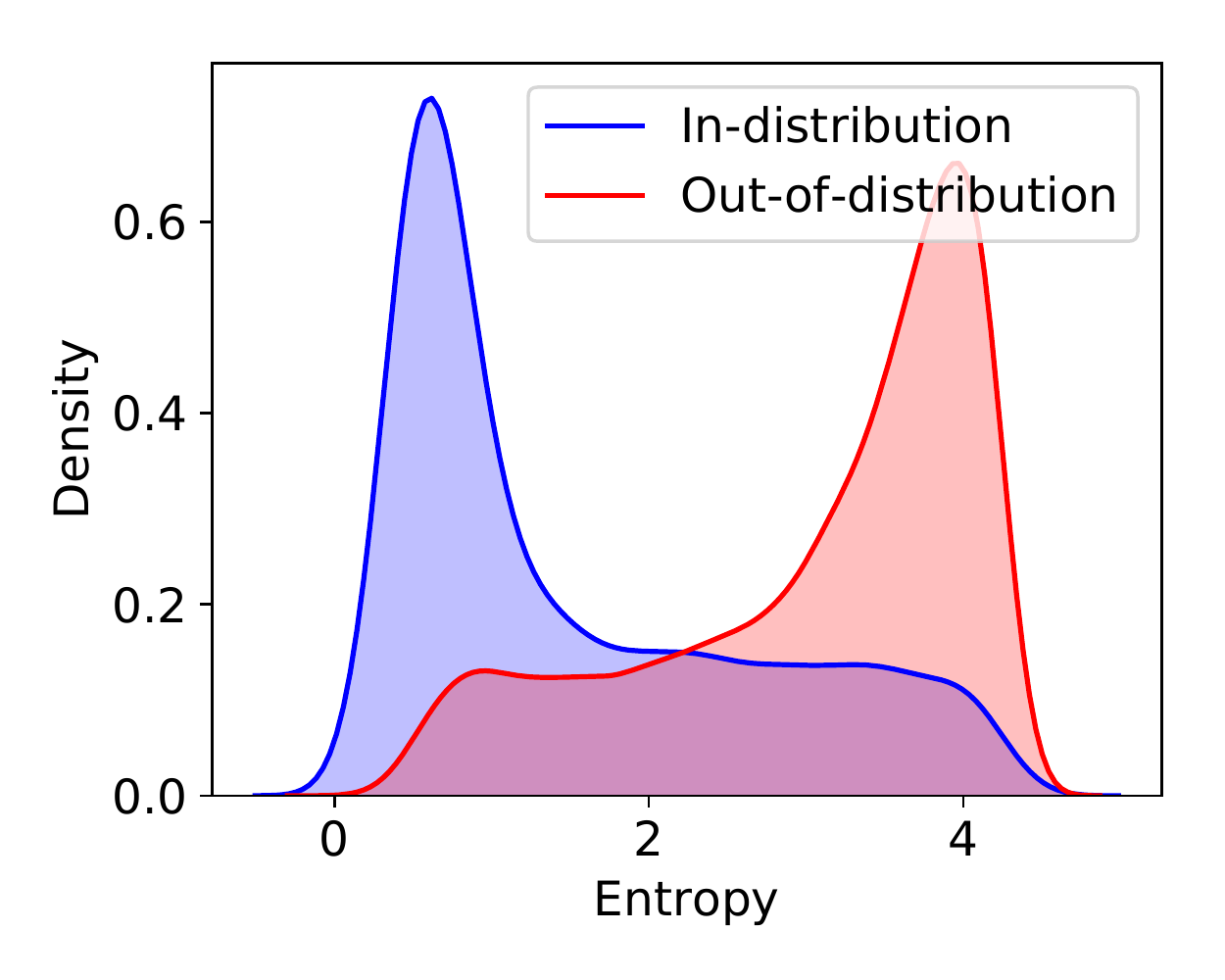}} 
    \end{subfigure} \\
    \begin{subfigure}[Deep ensemble \label{de_uncer}]
     {\includegraphics[width=0.23\textwidth]{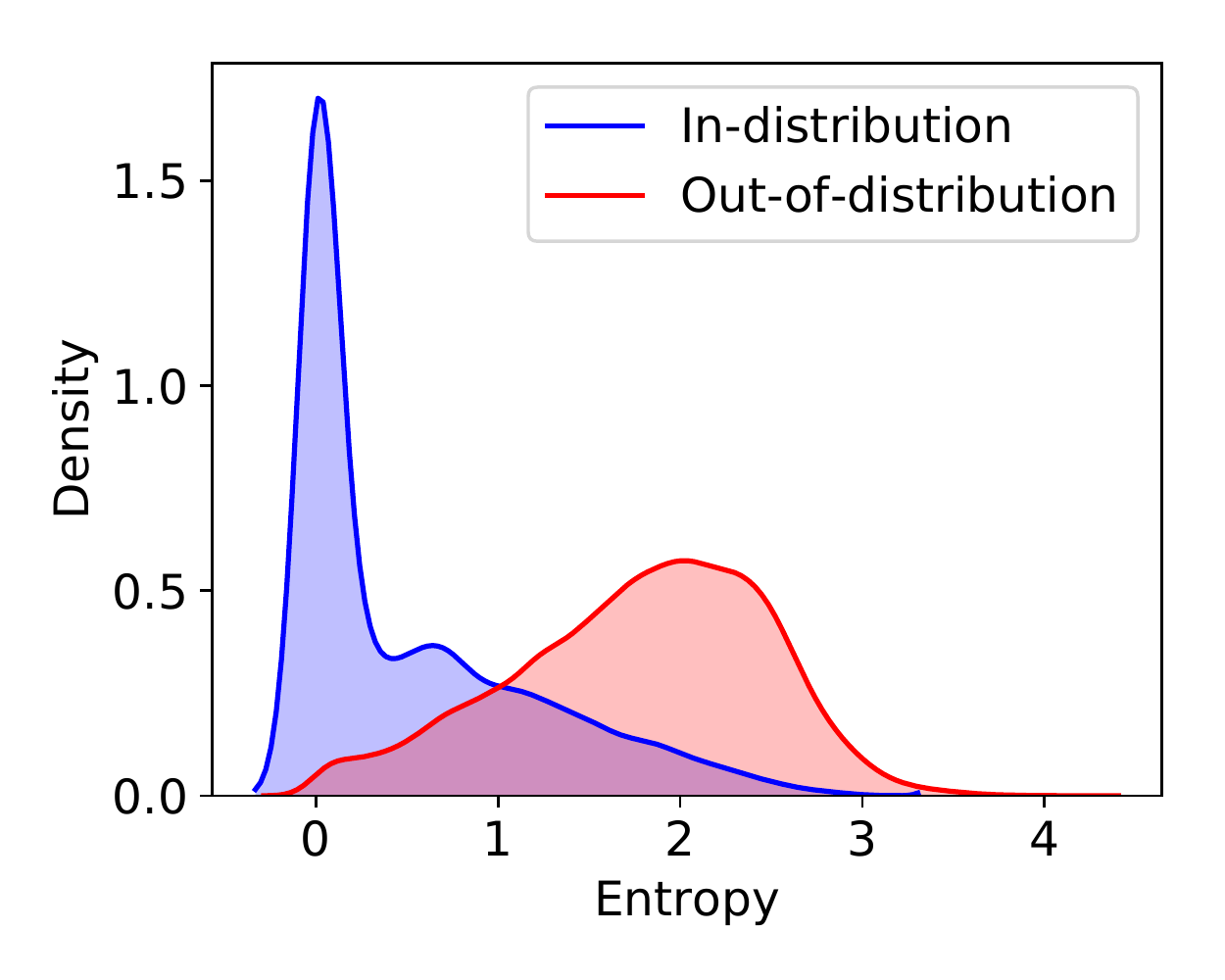}}
    \end{subfigure}
    \begin{subfigure}[MC dropout (all) \label{mc_uncer}]
     {\includegraphics[width=0.23\textwidth]{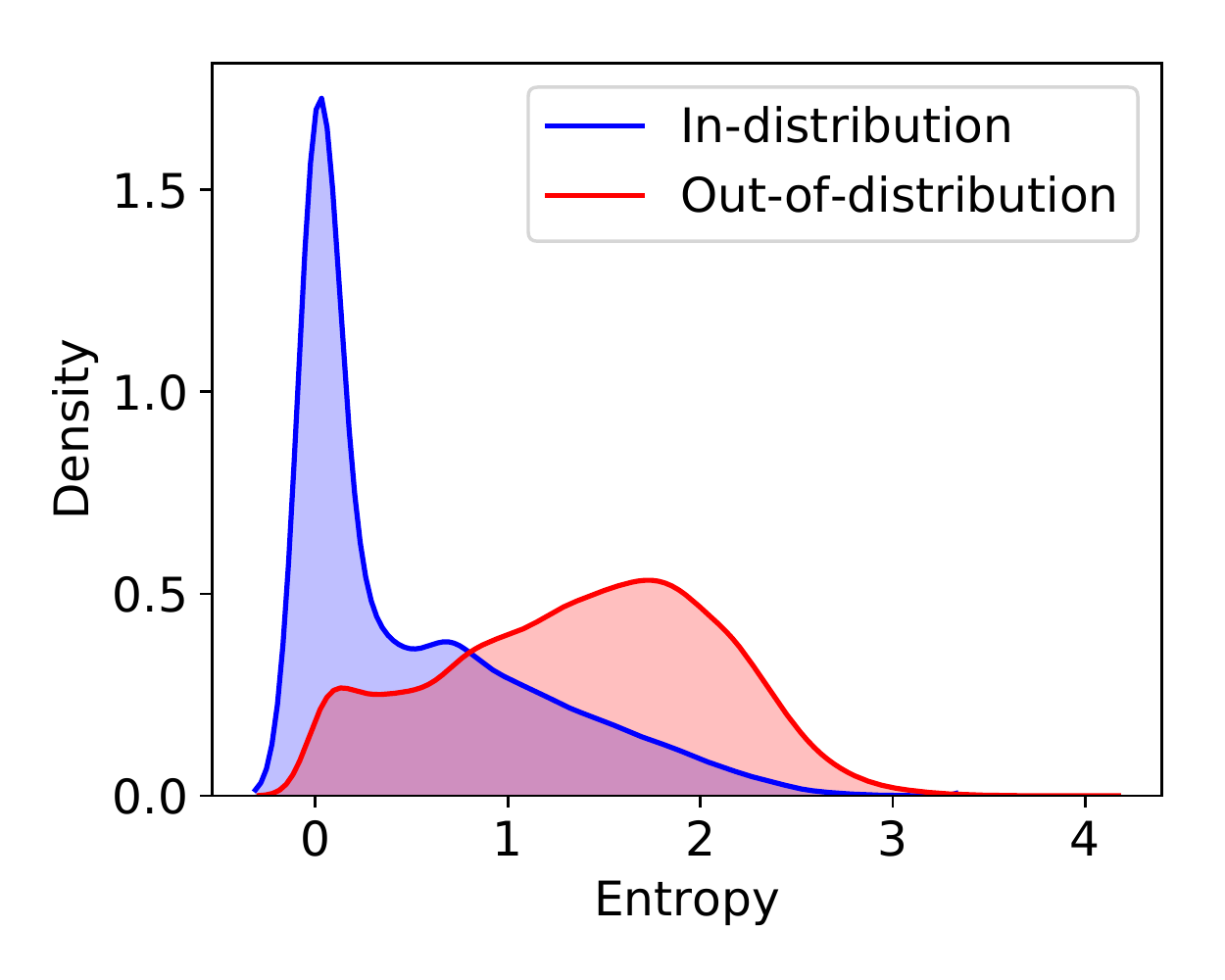}} 
    \end{subfigure}
    \caption{Uncertainty representation for in-distribution (CIFAR-100) and OOD (SVHN) of ResNet-50 under softmax and BM. 
    We exclude the result of last-layer MC dropout because it shows no meaningful difference compared to the softmax.
    }
    \label{uncertainty}
\end{figure}

\paragraph{Out-of-Distribution Uncertainty} 
We quantify uncertainty by predictive entropy, which measures the uncertainty of $f^{\mW}(\vx)$ as follows:
$H[\mathbb{E}_{q^{\mW}_{\rvz|\vx}}[\rvz]] = H[\phi(f^{\mW}(\vx))] = - \sum_{k=1}^K \phi_k (f^{\mW} (\vx)) \log \phi_k (f^{\mW} (\vx))$.
This uncertainty measure gives intuitive interpretation such that the ``I don't know" the answer is close to the uniform distribution; conversely, ``I confidently know" answer has one dominating categorical probability.

Figure~\ref{uncertainty} presents density plots of the predictive entropy, showing that BM provides notably better uncertainty estimation compared to other methods. 
Specifically, BM makes clear \textit{peaks} of predictive entropy in high uncertainty region for OOD samples (Figure~\ref{vc_uncer}). 
In contrast, softmax produces relatively flat uncertainty for OOD samples (Figure~\ref{sm_uncer}).
Even though both MC dropout and deep ensemble successfully increase predictive uncertainty for OOD samples compared to softmax, they fail to make a clear peak on the high uncertainty region for such samples, unlike BM.
We note that this remarkable result is obtained by being Bayesian only about the categorical probability.

Note that some in-distribution samples should be answered as ``I don't know" because the network does not achieve perfect test accuracy. 
As Figure~\ref{uncertainty} shows, BM contains more samples of high uncertainty for in-distribution samples compared to softmax that is almost certain in its predictions. 
This result consistently supports the previous result that BM resolves the overconfidence problem of softmax.

\begin{table}[t]
\caption{Transfer learning performance (test error rates) from ResNet-50 pretrained on ImageNet to smaller datasets. $\mu$ and $\sigma$ are obtained by five experiments, and boldface indicates the minimum mean error rate.
We examine only $\lambda = 0.01$ for BM.
}
\label{res50_transfer}
\vskip 0.1in
\begin{center}
\begin{small}
\begin{sc}
\begin{tabular}{lllll}
\toprule
Method      & C-10 &  Food-101 & Cars \\
\midrule
Softmax     & 5.44 $\pm 0.10$ &   28.49 $\pm 0.08$   &   42.99 $\pm 0.14$ \\
BM          & \textbf{5.03} $\pm 0.04$ &  \textbf{26.41}  $\pm 0.07$  &   \textbf{39.99} $\pm 0.20$ \\
\bottomrule
\end{tabular}
\end{sc}
\end{small}
\end{center}
\vskip -0.1in
\end{table}

\subsection{Transfer Learning} \label{sec:transfer_learning} 
BM adopts the Bayesian principle \textit{outside} the NNs, so it can be applied to models already trained on different tasks, unlike BNNs. 
In this regard, we examine the effectiveness of BM on the transfer learning scenario.
Specifically, we downloaded the ImageNet-pretrained ResNet-50, and fine-tune weights of the last linear layer for 100 epochs by the Adam optimizer \citep{kingma2015adam} with learning rate of 3e-4 on three different datasets (CIFAR-10, Food-101 \citep{bossard14}, and Cars \citep{krause20133d}.

\begin{figure}
    \centering
    \hfill
    \begin{subfigure}[Softmax]
     {\includegraphics[width=0.23\textwidth]{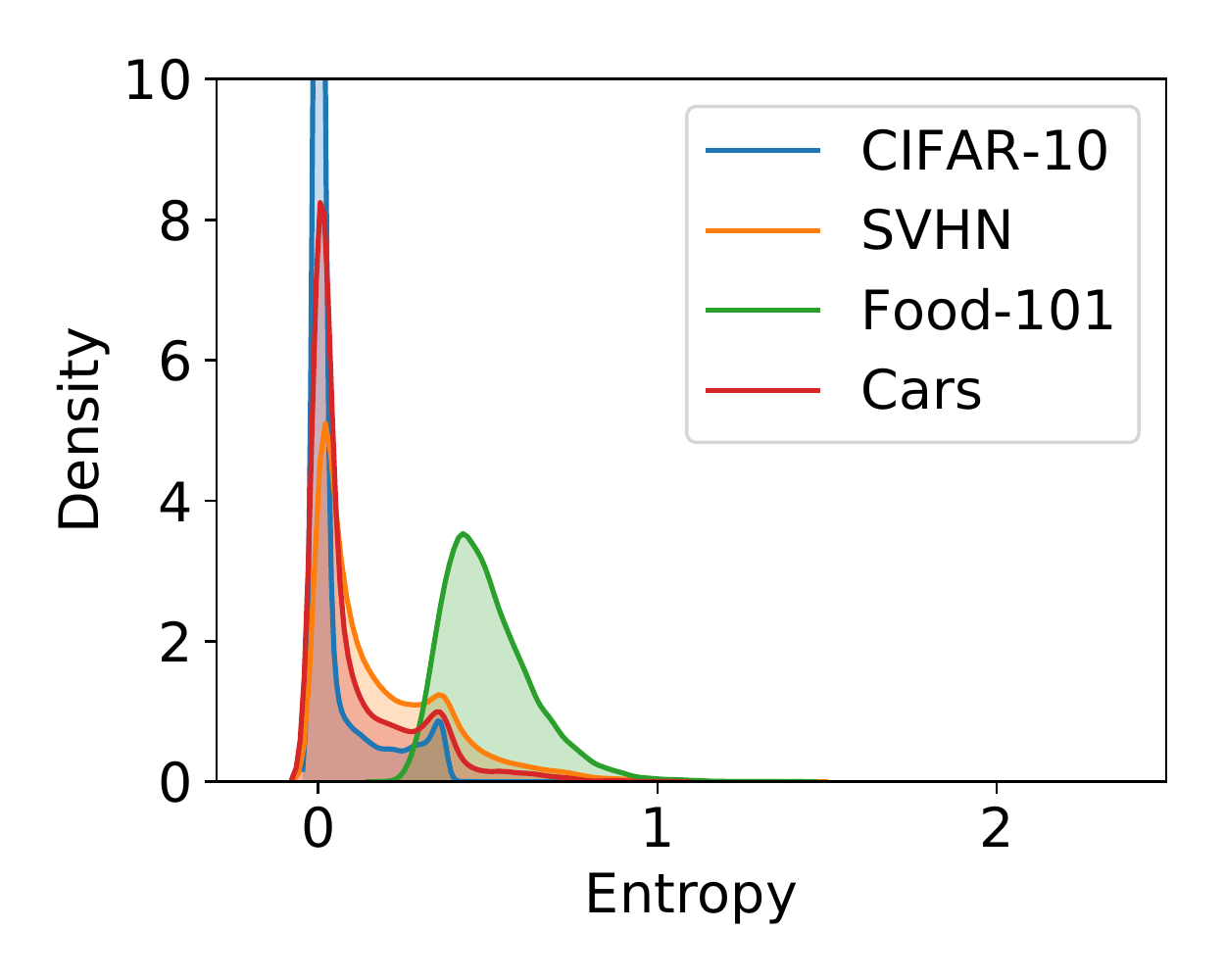}}
    \end{subfigure}
    \hfill
    \begin{subfigure}[BM]
     {\includegraphics[width=0.23\textwidth]{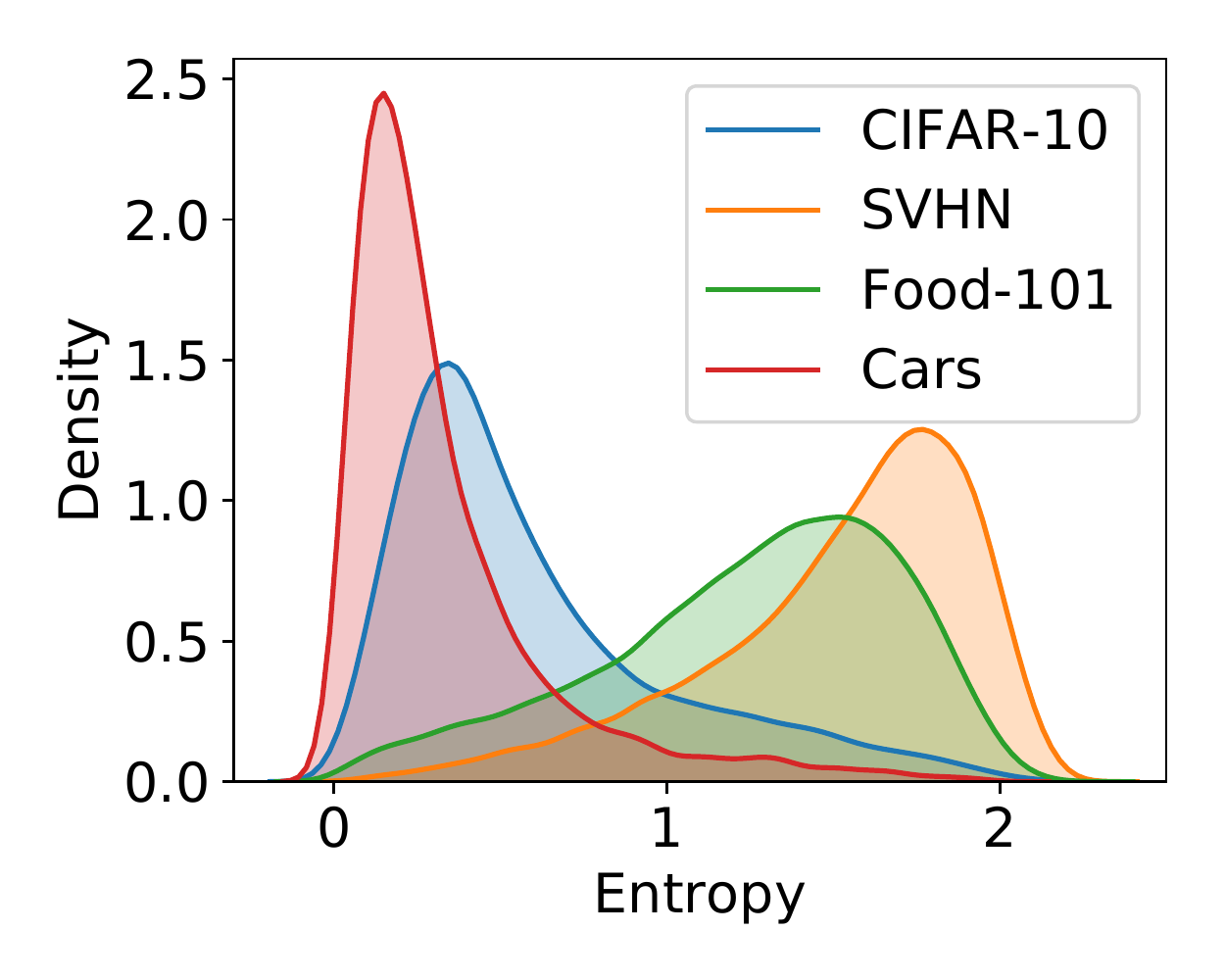}} 
    \end{subfigure}
    \hfill
    \caption{Uncertainty representation for in-distribution samples (CIFAR-10) and OOD samples (SVHN, Foods, and Cars) in transfer learning tasks.
    BM produces clear peaks in high uncertainty region on SVHN and Food-101. 
    We note that BM confidently predicts examples in Cars because CIFAR-10 contains the object category of ``automobile".  
    On the other hand, softmax produces confident predictions on all datasets compared to BM.}
    \label{transfer_uncertainty}
\end{figure}

Table~\ref{res50_transfer} compares test error rates of softmax and BM, in which BM consistently achieves better performances compared to softmax. 
Next, we examine the predictive uncertainty for OOD samples (Figure~\ref{transfer_uncertainty}). 
Surprisingly, we observe that BM significantly improves the uncertainty representation ability of pretrained-models \textit{by only fine-tuning the last layer weights}. 
These results present a possibility of adopting BM as a post-hoc solution to enhance the uncertainty representation ability of pretrained models without sacrificing their generalization performance. 
We believe that the interaction between BM and the pretrained models is significantly attractive, considering recent efforts of the deep learning community to construct general baseline models trained on extremely large-scale datasets and then transfer the baselines to multiple down-stream tasks (e.g., BERT \citep{devlin2018bert} and MoCo \citep{he2019momentum}).

\subsection{Semi-Supervised Learning} \label{sec:semi_supervised_learning}
BM enables NNs to represent rich information in their predictions (cf. section~\ref{sec:approx_dist}). 
We exploit this characteristic to benefit consistency-based loss functions for semi-supervised learning.
The idea of consistency-based losses employs information of unlabelled samples to determine where to promote robustness of predictions under stochastic perturbations \citep{belkin2006manifold,oliver2018realistic}.
In this section, we investigate two baselines that consider stochastic perturbations on inputs (VAT; \citeauthor{miyato2018virtual}, \citeyear{miyato2018virtual}) and networks ($\Pi$-model; \citeauthor{laine2016temporal}. \citeyear{laine2016temporal}), respectively.
Specifically, VAT generates adversarial direction $\vr$, then measures KL-divergence between predictions at $\vx$ and $\vx+\vr$:
\begin{equation}\label{eq:vat}
    \mathcal{L}_{VAT}(\vx) = KL (\phi(f^{\mW}(\vx)) \parallel \phi(f^{\mW}(\vx + \vr))) 
\end{equation}
where the adversarial direction is chosen by $\vr = \argmax_{\parallel \vr \parallel \leq \epsilon} KL (\phi(f^{\mW}(\vx)) \parallel \phi(f^{\mW}(\vx + \vr)))$, and $\Pi$-model measures $L^2$ distance between predictions with and without enabling stochastic parts in NNs:
\begin{equation}\label{eq:pi}
    \mathcal{L}_{\Pi}(\vx) = \parallel \phi(\bar{f}^{\mW}(\vx)) - \phi(f^{\mW}(\vx)) \parallel_2^2
\end{equation}
where $\bar{f}$ is a prediction without the stochastic parts.

We can see that both methods achieve the perturbation invariant predictions by minimizing the divergence between two categorical probabilities under some perturbations. 
In this regard, BM can provide a more delicate measure of the prediction consistency--divergence between Dirichlet distributions--that can capture richer probabilistic structures, e.g., (co)variances of categorical probabilities.
This generalization the moment matching problem to the distribution matching problem can be achieved by replacing only the consistency measures in \eqref{eq:vat} with $KL (q_{\rvz|\vx}^{\mW} \parallel q_{\rvz|\vx + \vr }^{\mW} )$ and in \eqref{eq:pi} with $KL (\bar{q}_{\rvz|\vx}^{\mW} \parallel q_{\rvz|\vx }^{\mW} )$.

We train wide ResNet 28-2 \citep{zagoruyko2016wide} via the consistency-based loss functions on CIFAR-10 with 4K/41K/5K number of labeled training/unlabeled training/validation samples.
Our experimental results show that the distribution matching metric of BM is more effective than the moment matching metric under the softmax on reducing the error rates (Table~\ref{semi_sup}). 
We think that the improvement in semi-supervised learning with more sophisticated consistency measures shows the potential usefulness of BM on other useful applications.
This is because employing the prediction difference of neural networks is a prevalent method in many domains such as knowledge distillation \citep{hinton2015distilling} and model interpretation \citep{zintgraf2017visualizing}.

\begin{table}[t]
\caption{Classification error rates on CIFAR-10. $\mu$ and $\sigma$ are obtained by five experiments, and boldface indicates the minimum mean error rate.
We matched the configurations to those of \citet{oliver2018realistic} except for a consistency loss coefficient of 0.03 for VAT and 0.5 for $\Pi$-model to match the scale between supervised and unsupervised losses. 
We use only $\lambda= 0.01$ for BM.
}
\label{semi_sup}
\vskip 0.1in
\begin{center}
\begin{small}
\begin{sc}
\begin{tabular}{lll}
\toprule
Method   & $\Pi$-Model & VAT \\
\midrule
Softmax     & 16.52 $\pm 0.21$     & 13.33 $\pm 0.37$    \\
BM          & \textbf{16.01} $\pm 0.36$ & \textbf{12.40} $\pm 0.23$    \\
\bottomrule
\end{tabular}
\end{sc}
\end{small}
\end{center}
\vskip -0.1in
\end{table}

\section{Conclusion}
We adopted the Bayesian principle for constructing the target distribution by considering the categorical probability as a random variable rather than being given by the training label.
The proposed method can be flexibly applied to the standard deep learning models by replacing only the softmax and the cross-entropy loss, which provides the consistent improvements in generalization performance, better uncertainty estimation, and well-calibrated behavior.
We believe that BM shows promising advantages of being Bayesian about categorical probability.

We think that accommodating more expressive distributions in the belief matching framework is an interesting future direction.
For example, parameterizing the logistic normal distribution (or mixture distribution) can make neural networks to capture strong semantic similarities among class labels, which would be helpful in large-class classification problems such as machine translation and classification on the ImageNet.
Besides, considering the input dependent prior would result in interesting properties. 
For example, under the teacher-student framework, the teacher can dictate the prior for each input, thereby control the desired smoothness of the student's prediction on the location. 
This property can benefit various domains such as imbalanced datasets and multi-domain learning.


\section*{Acknowledgements}
We would like to thank Dong-Hyun Lee and anonymous reviewers for the discussions and suggestions.

\bibliography{icml2020}
\bibliographystyle{icml2020}

\end{document}